\documentclass[journal]{IEEEtran}
\ifCLASSINFOpdf
\else
\fi

\usepackage{times}
\usepackage{epsfig}
\usepackage{graphicx}
\usepackage{amsmath}
\usepackage{amssymb}
\usepackage{multirow}
\usepackage{xcolor}

\begin{document}
%
\title{Convolutional Neural Networks with Dynamic Regularization}
%
%
%

\author{Yi~Wang, \textit{Student Member, IEEE}, Zhen-Peng~Bian, Junhui~Hou, \textit{Senior Member, IEEE}, \\and Lap-Pui~Chau, \textit{Fellow, IEEE}
\thanks{Yi~Wang and Lap-Pui~Chau are with School of  Electrical and Electronics Engineering, Nanyang Technological University, Singapore, 639798 (e-mail: wang1241@e.ntu.edu.sg, elpchau@ntu.edu.sg).}
\thanks{Zhen-Peng~Bian is with Singapore Telecommunications Limited, Singapore, 239732 (e-mail: zbian1@ntu.edu.sg).}
\thanks{Junhui~Hou is with Department of Computer Science, City University of Hong Kong  (e-mail: jh.hou@cityu.edu.hk).}
\thanks{Yi~Wang and Zhen-Peng~Bian contributed to this work equally.}
\thanks{Corresponding author: Lap-Pui Chau.}
\thanks{This work was supported in part by the Hong Kong Research Grants Council under grants 9048123 (CityU 21211518) and 9042820 (CityU 11219019).}
}

%
%

\markboth{Journal of \LaTeX\ Class Files}%
{Shell \MakeLowercase{\textit{et al.}}: Bare Demo of IEEEtran.cls for IEEE Journals}
%



\maketitle

\begin{abstract}
Regularization is commonly used for alleviating overfitting in machine learning. For convolutional neural networks (CNNs), regularization methods, such as DropBlock and Shake-Shake, have illustrated the improvement in the generalization performance. However, these methods lack a self-adaptive ability throughout training. That is, the regularization strength is fixed to a predefined schedule, and manual adjustments are required to adapt to various network architectures. In this paper, we propose a dynamic regularization method for CNNs. Specifically, we model the regularization strength as a function of the training loss. According to the change of the training loss, our method can dynamically adjust the regularization strength in the training procedure, thereby balancing the underfitting and overfitting of CNNs. With dynamic regularization, a large-scale model is automatically regularized by the strong perturbation, and vice versa. Experimental results show that the proposed method can improve the generalization capability on off-the-shelf network architectures and outperform state-of-the-art regularization methods.
\end{abstract}


%
\IEEEpeerreviewmaketitle

\section{Introduction}

\IEEEPARstart{C}{onvolutional} neural networks (CNNs), which use a stack of convolution operations followed by non-linear activation (e.g., Rectified Linear Unit, ReLU) to extract high-level discriminative features, have achieved considerable improvements for visual tasks \cite{krizhevsky2012imagenet,he2016deep,zhao2019object}. Recent advances of the CNN architectures, such as ResNet \cite{he2016deep}, DenseNet \cite{huang2017densely}, ResNeXt \cite{xie2017aggregated}, and PyramidNet \cite{han2017deep}, ease the vanishing gradient problem and boost the performance. However, CNNs still suffer from the overfitting problem, which reduces their generalization capability.

A wide variety of regularization strategies were exploited to alleviate overfitting and decrease the generalization error. Data augmentation \cite{krizhevsky2012imagenet} is a simple yet effective manner to improve the diversity of training data. Batch normalization \cite{ioffe2015batch} standardizes the mean and variance of features of each mini-batch, which makes the optimization landscape smoother \cite{santurkar2018does}. Dropout \cite{srivastava2014dropout} aims to train an ensemble of sub-networks, weakening the effect of ``co-adaptions'' on training data. DropBlock \cite{ghiasi2018dropblock} introduces a structured dropout approach, which drops the contiguous regions of a feature map. Shake-Shake regularization \cite{gastaldi2017shake} was proposed to randomly interpolate two complementary features in the two residual branches of ResNeXt, achieving state-of-the-art classification performance. ShakeDrop \cite{yamada2018shakedrop} incorporates the idea of stochastic depth \cite{huang2016deep} with Shake-Shake regularization to stabilize the training process for ResNet-like architectures. Despite the impressive improvement of the regularization methods, there are two main drawbacks with these methods.
\begin{enumerate}
\item The regularization strength (or amplitude) is not flexible for different network architectures. For example, the ShakeDrop was designed for deep networks but not suitable for shallow networks. Instead of improving classification performance, it even worsens the performance of shallow networks (see the TABLE \ref{table1}).
\item The regularization strength is unchangeable over the whole training process. The fixed strong regularization is beneficial to reduce overfitting, but it causes difficulties to fit data at the beginning of training. From the perspective of curriculum learning \cite{bengio2009curriculum}, the learner should begin with easy examples.
\end{enumerate}

In view of these issues, we propose a dynamic regularization method for CNNs, in which the regularization strength is adaptable to the change of the training loss. During training, the regularization strength is gradually increased with respect to the training status. Analogous to human education, the regularizer is regarded as an instructor who gradually increases the difficulty of training examples in a form of feature perturbation. The dynamic regularization can adapt to different model sizes. It provides a strong regularization for large-scale models, and vice versa. (See Fig. \ref{Fig5} (b)). That is, the regularization strength grows faster and achieves a higher value for a large-scale model than that of a light model.

\begin{figure*} [ht]
\centering 
\includegraphics[width=13cm]{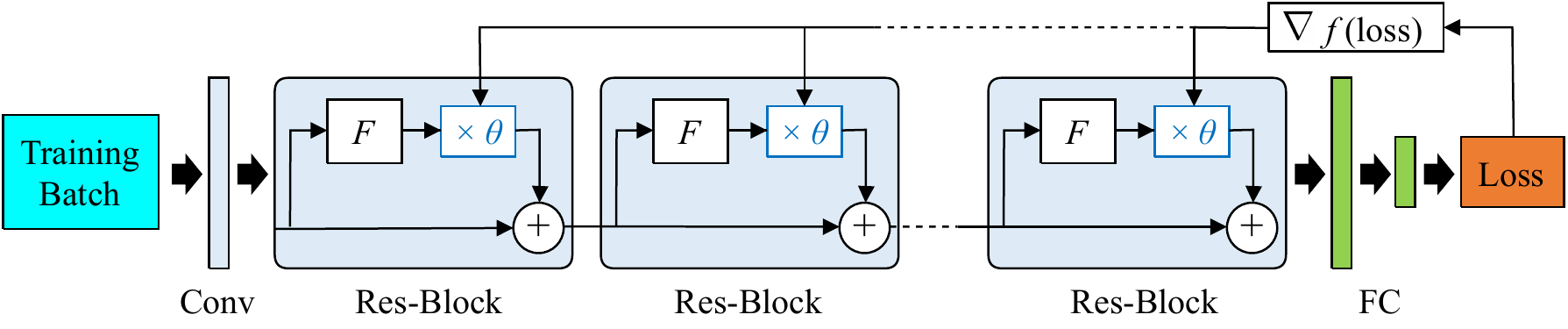}\\ 
\caption{The proposed dynamic regularization in the ResNet structure. Conv denotes the convolutional layer. FC denotes the fully connected layer. \(F\) denotes the residual function. \(\nabla f(loss)\) denotes a backward difference of the training loss. The dynamic regularization aims to make a self-adaptive schedule throughout training for various network sizes by adjusting the strength of the random perturbation \(\theta\). As a manner of feature augmentation, the \(\theta\) introduces noises for the residual branch in the forward and backward process.}\label{Fig1} 
\end{figure*}

Fig. \ref{Fig1} shows the proposed dynamic regularization in the ResNet structure. The training loss is not only used to perform backpropagation but also exploited to update the amplitude of the regularization. The features are multiplied by the regularizer in the residual branch. The regularizer works as a perturbation which introduces an augmentation in feature space, so CNNs are trained by the diversity of augmented features. Additionally, the regularization amplitude is changeable with respect to the change of the training loss. We conduct experiments on the image classification task to evaluate our regularization strategy. Experimental results show that the proposed dynamic regularization outperforms state-of-the-art regularization methods, i.e., PyramidNet, ResNeXt, and DenseNet equipped with our dynamic regularization improve the classification accuracy in various model settings, when compared with the same networks with ShakeDrop \cite{yamada2018shakedrop}, Shake-Shake \cite{gastaldi2017shake}, and DropBlock \cite{ghiasi2018dropblock}, respectively.

The rest of this paper is organized as follows. We first briefly introduce the related work on deep CNNs and regularization methods in Section \ref{Related}. Then, the proposed dynamic regularization is presented in Section \ref{Dynamic}. Experimental results and discussion are given in Section \ref{Experiment}. Finally, Section \ref{Conclusion} concludes this paper.

\section{Related Work} \label{Related}

\subsection{Deep CNNs}
CNNs have become deeper and wider with a more powerful capacity \cite{he2016deep,huang2017densely,han2017deep,simonyan2014very,szegedy2015going}. As our proposed regularization is based on ResNet and its variants, we briefly review the basic structure of ResNet, i.e., the residual block.

\textbf{Residual block.} The residual block (Res-Block, shown in Fig. \ref{Fig1}) is formulated as
\begin{equation}
\label{eqn1}
x_{l+1} = x_{l} + F(x_{l},\mathcal{W}_{l}),
\end{equation}
where an identity branch \(x_{l}\) is the input features of the \(l^{th}\) Res-Block, which is added by a residual branch \(F\) that is a non-linear transformation between \(x_{l}\) and a set of parameters \(\mathcal{W}_{l}\) (\(\mathcal{W}_{l}\) will be omitted for simplicity in the following). \(F\) consists of two Conv-BN-ReLU or Bottleneck Architectures in the original ResNet structure \cite{he2016deep}. In recent works, \(F\) was also designed to other forms, e.g. Wide-ResNet \cite{zagoruyko2016wide}, Inception module \cite{szegedy2017inception}, PyramidNet \cite{han2017deep}, and ResNeXt \cite{xie2017aggregated}. PyramidNet gradually increases the number of channels in the Res-Blocks as the layers go deep. ResNeXt has multiple aggregated residual branches expressed as
\begin{equation}
\label{eqn2}
x_{l+1} = x_{l} + F_{1}(x_{l}) +F_{2}(x_{l}),
\end{equation}
where \(F_{1}\) and \(F_{2}\) are two residual branches. The number of branches (namely cardinality) is not limited.

\begin{figure*} [ht]
\centering 
\includegraphics[width=16.2cm]{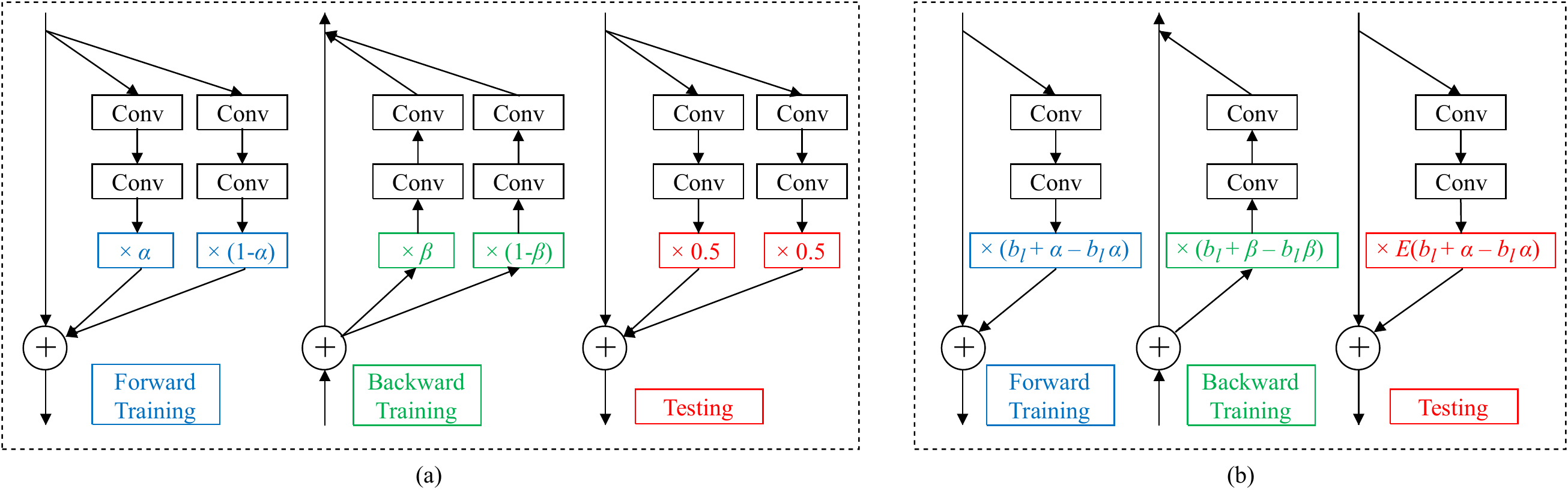}\\ 
\caption{Shake-based regularization methods in the Res-Block. Some layers (e.g., batch normalization and ReLU) in the residual branch is omitted for simplicity. (a) 3-branch architecture with Shake-Shake regularization \cite{gastaldi2017shake}. (b) 2-branch architecture with ShakeDrop \cite{yamada2018shakedrop}.}\label{Fig2} 
\end{figure*}

\subsection{Regularization}
In addition to the advances of network architectures, many regularization techniques, e.g., data augmentation \cite{krizhevsky2012imagenet,devries2017improved}, stochastic dropping \cite{srivastava2014dropout,huang2016deep,larsson2016fractalnet,morerio2017curriculum,ghiasi2018dropblock}, and Shake-based regularization methods \cite{gastaldi2017shake,yamada2018shakedrop}, have been successfully applied to avoid overfitting of CNNs.

Data augmentation (e.g., random cropping, flipping, and color adjusting \cite{krizhevsky2012imagenet}) is a simple yet effective strategy to increase the diversity of data. DeVries and Taylor \cite{devries2017improved} introduced an image augmentation technique, where augmented images are generated by randomly cutting out square regions from input images (called Cutout). Dropout \cite{srivastava2014dropout} is a widely-used technique which stochastically drops out the hidden nodes from the networks during the training process. Following this idea, Maxout \cite{goodfellow2013maxout}, Continuous Dropout \cite{shen2017continuous}, DropPath \cite{larsson2016fractalnet}, and stochastic depth \cite{huang2016deep} were proposed. Stochastic depth randomly drops a certain number of residual branches of ResNet so that the network is shrunk in training. By incorporating Dropout with Cutout, DropBlock \cite{ghiasi2018dropblock} drops the contiguous regions in a feature map. Adding a parameter norm penalty to the loss function, the weight decay (or Tikhonov regularization) is commonly used for neural networks and linear inverse problems \cite{de2003regularization}. DisturbLabel \cite{xie2016disturblabel} imposes noisy labels in the loss function. Shake-based regularization approaches \cite{gastaldi2017shake,yamada2018shakedrop} were recently proposed to augment features inside CNNs, which achieve appealing classification performance. 

\textbf{Shake-based regularization approaches.} Gastaldi \cite{gastaldi2017shake} proposed a Shake-Shake regularization method, as shown in Fig. \ref{Fig2} (a). A random variable \(\alpha\) is used to control the interpolation of the two residual branches (i.e., \(F_{1}(x)\) and \(F_{2}(x)\) in 3-branch ResNeXt). It is given by:
\begin{equation}
\label{eqn3}
x_{l+1} = x_{l} + \alpha F_{1}(x_{l}) +(1-\alpha)F_{2}(x_{l}),
\end{equation}
where \(\alpha \in [0,1]\) follows the uniform distribution in the forward pass. For the backward pass, \(\alpha\) is replaced by another uniform random variable \(\beta \in [0,1]\) to disturb the learning process. The regularization amplitude of each branch is fixed to \(1\).

To extend the use of Shake-Shake regularization, Yamada \textit{et al.} \cite{yamada2018shakedrop} introduced a single Shake in 2-branch architectures (e.g., ResNet or PyramidNet) as shown in Fig. \ref{Fig2} (b). Stochastic depth \cite{huang2016deep} was adopted to stabilize the learning: 
\begin{equation}
\label{eqn4}
x_{l+1} = x_{l} + (b_{l}+\alpha-b_{l}\alpha)F(x_{l}),
\end{equation}
where \(\alpha \in [-1,1]\) is a uniform random variable and \(b_{l} \in \left \{ 0,1 \right \}\) is the Bernoulli random variable determining when to perform the original network (i.e., \(x_{l+1} = x_{l} + F(x_{l})\), if \(b_{l}=1\)) or the perturbated one (i.e., \(x_{l+1} = x_{l} + \alpha F(x_{l})\), if \(b_{l}=0\)). In the backward pass, \(\alpha\) is replaced by \(\beta \in [0,1]\). The probability of \(b_{l}\) denotes \(p_{l}=P(b_{l}=1)\), which follows a  linear decay rule, i.e., \(p_{l}=1-\frac{l}{L}(1-p_{L})\), where \(L\) is the total number of Res-Blocks and \(p_{L}=0.5\). The regularization amplitude of the branch is also fixed to \(1\). We argue that this heavy regularization overemphasizes the overfitting and the fixed regularization amplitude cannot fit the dynamics of the training process and different model sizes well. 

\section{The Proposed Method}\label{Dynamic} 

As aforementioned, the fixed regularization strength in the existing regularization methods, such as DropPath \cite{larsson2016fractalnet}, Stochastic depth \cite{huang2016deep}, Shake-Shake \cite{gastaldi2017shake}, and Shakedrop \cite{yamada2018shakedrop}, departs from the human learning paradigm (e.g., the curriculum learning \cite{bengio2009curriculum,morerio2017curriculum} or self-paced learning \cite{kumar2010self}). A naive way is to predefine the schedule for updating the regularization strength, such as the linear increment scheme in \cite{ghiasi2018dropblock,zoph2018learning}, which linearly increases the regularization strength from low to high. We argue that the predefined schedule is not flexible enough to reveal the learning process. Based on the fact that the loss of the learning system can fully provide the learning status, we propose a dynamic regularization, which is capable of adjusting the regularization strength adaptively.

Our dynamic regularization for CNNs leverages the dynamics of the training loss. That is, at the beginning of training, both the training and testing losses keep decreasing. Through a certain number of iterations, the network overfits the training data, resulting in that the training loss decreases more rapidly than the testing loss. We design a regularization strategy to follow this dynamics. If the training loss drops in an iteration, the regularization strength should increase against the overfitting in the next iteration; otherwise, the regularization strength should decrease against the underfitting. In what follows, we first introduce the dynamic regularization in the residual architectures and then deliberate the update of the regularization strength in each iteration of the training process. We finally extend our dynamic regularization in the densely-connected networks.

\begin{figure} [t]
\centering 
\includegraphics[width=7.5cm]{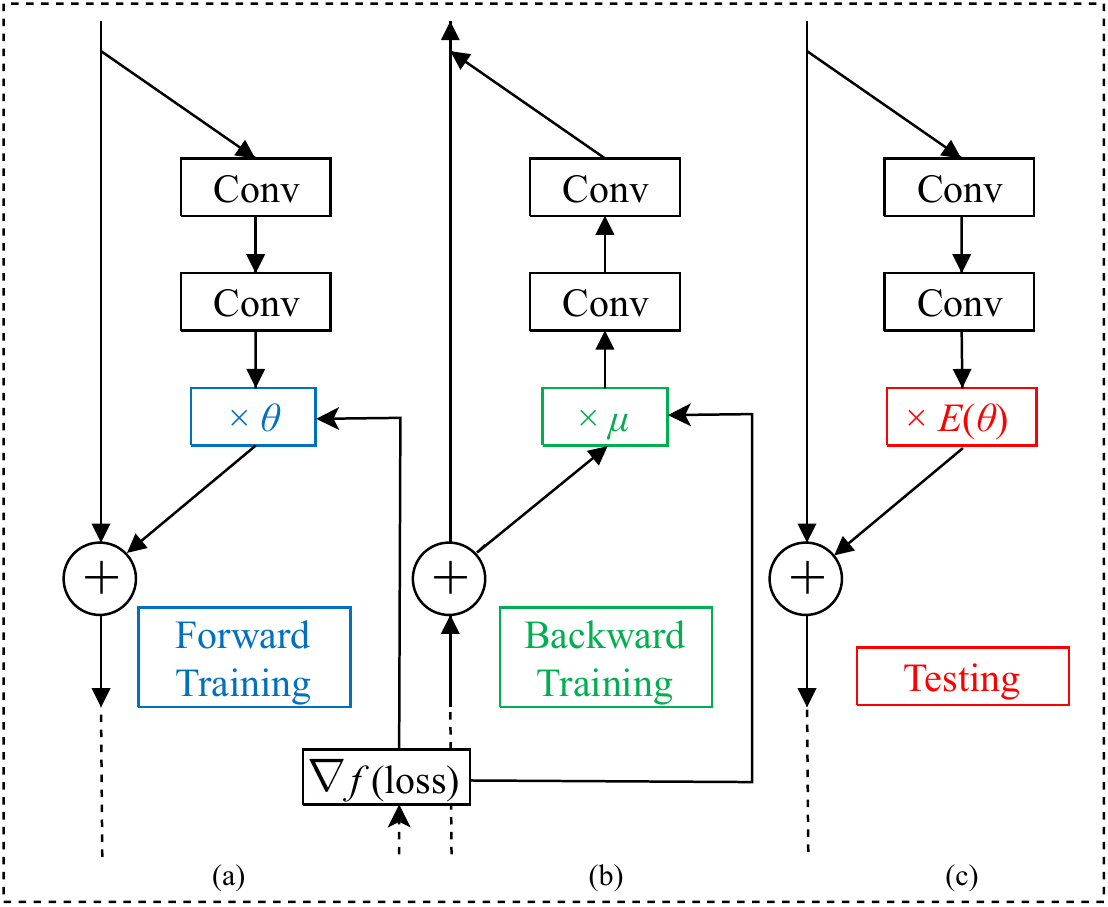}\\ 
\caption{The 2-branch Res-Block with dynamic regularization.}\label{Fig3} 
\end{figure}

\subsection{Residual Architectures with Dynamic Regularization}
We apply the dynamic regularization method in two residual network architectures: the 2-branch architecture (e.g., PyramidNet \cite{han2017deep}) and the 3-branch architecture (e.g., ResNeXt \cite{xie2017aggregated}).

\subsubsection{The 2-branch architecture with dynamic regularization}
\textbf{Training phase.} The dynamic regularization adopted in a Res-Block is shown in Figs. \ref{Fig3} (a) and (b). Specifically, a dynamic regularization unit (called random perturbation) is embedded into the residual branch of a Res-Block. The random perturbation \(\theta\) is achieved by 
\begin{equation} \label{eqn5} 
\theta = A + s_{i}\cdot r,
\end{equation}
where \(A\) is the basic constant amplitude, \(s_{i}\) is the dynamic factor at the \(i^{th}\) iteration, and \(r \in [-R,R]\) is the uniform random noise with the expected value \(E(r)=0\). The value of $s_i$ is updated via the backward difference of the training loss (See Section \ref{Dynamic}.B). The regularization amplitude is proportional to \(A+s_{i}\cdot R\). In the forward pass, the output of the \((l+1)^{th}\) Res-Block can be expressed as:
\begin{equation}
\label{eqn6}
x_{l+1} = x_{l} + (A+s_{i}\cdot r)F(x_{l}).
\end{equation}
In the backward pass, \(\theta\) has a different value (represented by \(\mu\) in Fig. \ref{Fig3} (b)) due to the random noise \(r\).

\textbf{Random noise.} The range of \(r\), i.e., $R$, is a hyper-parameter in the training phase. A straightforward way is to set \(R\) to be uniform inside all Res-Blocks. According to \cite{huang2016deep}, the features of the bottom Res-Blocks should remain more than those of the top Res-Blocks. Hence, we propose a linear enhancement rule to configure this range inside Res-Blocks. For the \(l^{th}\) Res-Block, the range denoted as $R_l$ is given by
\begin{equation}
\label{eqn7}
R_{l} = l/L,
\end{equation}
where \(L\) is the total number of Res-Blocks. With the linearly increased \(R\), the regularization strength is gradually raised from the bottom layers to the top layers. We conducted comparative experiments on different settings of \(R\) in Section \ref{Experiment}.C.3.

\textbf{Inference phase.} As shown in Fig. \ref{Fig3} (c), we calculate the expected value of \(\theta\) as
\begin{equation}
\label{eqn8}
E(\theta) = E(A+s_{i}\cdot r)=A,
\end{equation}
and obtain a forward pass for inference:  
\begin{equation}
\label{eqn9}
x_{l+1} = x_{l} +A\cdot F(x_{l}).
\end{equation}
Since \(A\) is a constant, Eq. (\ref{eqn9}) is equivalent to the standard Res-Block. 

\subsubsection{The 3-branch architecture with dynamic regularization}
As shown in Fig. \ref{Fig2} (a), we apply the dynamic regularization in the 3-branch architecture. Formally, we use the proposed random perturbation \(\theta\) of Eq. (\ref{eqn5}) to replace \(\alpha\) of Eq. (\ref{eqn3}) in Shake-Shake regularization. Hence, the Res-Block with dynamic regularization can be defined as
\begin{equation}
\label{eqn10}
x_{l+1} = x_{l} + (A+s_{i}\cdot r)F_{1}(x_{l}) +(1-A-s_{i}\cdot r)F_{2}(x_{l}).
\end{equation}
If we set \(A=0.5\), \(r \in [-0.5, 0.5]\), and \(s_{i}=1\), then \(\theta\) ranges from \(0\) and \(1\), which is equivalent to \(\alpha\) of Eq. (\ref{eqn3}). The Shake-Shake regularization can be thought of as a special case of our dynamic regularization with a fixed strength.

\subsection{Update of the Regularization Strength}

The proposed updating solution for the dynamic regularization strength is achieved by the dynamics of the training loss. In particular, the dynamic characteristic of the training loss can be model as the backward difference between the training losses at successive iterations: 
\begin{equation}
\label{eqn11}
\nabla loss_{i}= loss_{i} - loss_{i-1},
\end{equation}
where \(loss_{i}\) denotes the training loss at the \(i^{th}\) iteration. Although the training loss shows a downtrend in training, large fluctuations appear when sequential mini-batches are fed. To eliminate the fluctuations and obtain the overall trend of the loss, we apply a Gaussian filter to smooth it. The filtered backward difference can be rewritten as
\begin{equation}
\label{eqn12}
\nabla f(loss_{i})= f(loss_{i}) - f(loss_{i-1}),
\end{equation}
where \(f(\cdot )\) is the filtering operation defined as 
\begin{equation}
\label{eqn13}
f(loss_{i})=\sum_{n=0}^{N}w[n]\cdot loss_{i-n}.
\end{equation}
The filter length is \(N+1\). Here we use the normalized Gaussian window and formulate \(w[n]\) as 
\begin{equation}
\label{eqn14}
w[n]=\frac{1}{\sqrt{2\pi}(\sigma N/2)}e^{-\frac{1}{2}\left ( \frac{n-N/2}{\sigma N/2} \right )^{2}},
\end{equation}
where \(\sigma=0.4\), and \(0\leq n \leq N\). The standard deviation is determined by \(\sigma \cdot N/2\). We will discuss the effectiveness of the Gaussian filter in Section \ref{Experiment}.C.4. The dynamic factor in Eqs. (\ref{eqn6}) and (\ref{eqn10}) is updated with respect to \(\nabla f(loss_{i})\), i.e., 
\begin{equation}
\label{eqn15}
s_{i+1} = \left\{\begin{matrix}
s_{i} + \Delta s, & \nabla f(loss_{i})\leq  0\\ 
s_{i} - \Delta s, & \nabla f(loss_{i})>  0
\end{matrix}\right.
\end{equation}
where \(\Delta s\) is a small constant step for changing the regularization amplitude. From Eq. (15), it can be observed that if the training loss decreases (\(\nabla f(loss_{i})\leq  0\)), the regularization amplitude increases to avoid overfitting; otherwise, it decreases to prevent underfitting. The dynamic factor keeps updating to reflect the dynamics of the training loss. 

\textbf{Remark.} Some methods have been proposed to change the regularization strength. For instance, Zoph \textit{et al.} \cite{zoph2018learning} introduced a linear increment scheme, ScheduledDropPath, to regularize NASNets. The probability of dropping out a path is increased linearly throughout training. Following this, DropBlock \cite{ghiasi2018dropblock} employs a linearly-increased dropping rate. However, the constant or linear scheme is still a predefined rule, which cannot adapt to the training procedure and different model size. Different from them, our dynamic scheduling exploits the dynamics of the training loss, which is applicable to different network architectures. In Section \ref{Experiment}.C.2, we conducted comparisons between them.

\begin{figure} [t]
\centering 
\includegraphics[width=7.5cm]{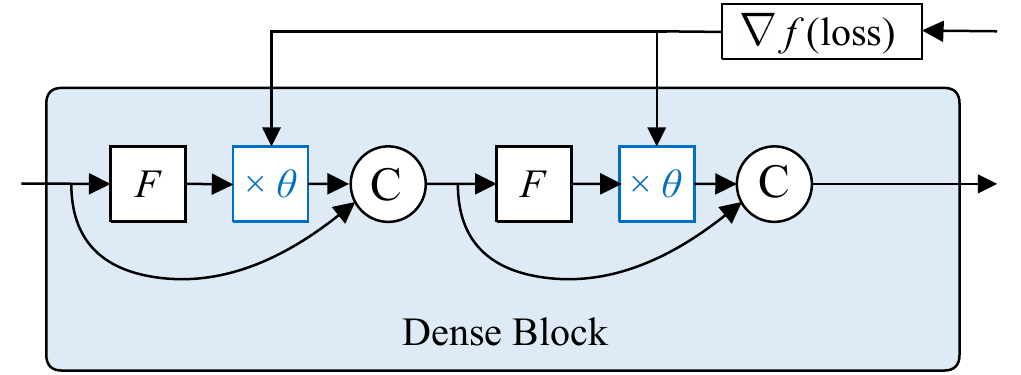}\\ 
\caption{Dense block with dynamic regularization. \(F\) denotes a convolution operation. \(\theta\) is a random perturbation. `C' means a concatenation operation. \(\nabla f(loss)\) denotes a backward difference of the training loss.}\label{Fig4} 
\end{figure}

\subsection{Extension to Densely-Connected Networks}

To illustrate the flexibility of our method, we further incorporate it into DenseNet \cite{huang2017densely} by assigning  the random perturbations inside the dense block. Fig. \ref{Fig4} shows a two-layer dense block with dynamic regularization, where the perturbations are inserted behind the output features of the convolutional layers. This manner accumulates the noise from all preceding layers to the current layer. A small perturbation could lead to serious noise for the subsequent layers. In experiments, we found that ShakeDrop and DropBlock with default hyper-parameters yield worse results. This is caused by the strong regularization. To decrease the regularization strength, we increase the probability of the Bernoulli random variable (i.e., \(p_{L}\), the rate of keeping original features rather than shaking features) in Eq. (\ref{eqn4}) for ShakeDrop and increase the \textit{keep\_prob} for DropBlock. Note that for our dynamic regularization, it is not needed to adjust the hyper-parameters used in the residual structure. Our method obtains consistent better results. More details can be seen in Section \ref{Experiment}.

\begin{table}[t]
\centering \caption{Comparison of regularization methods on CIFAR100 in the 2-branch architecture (i.e., PyramidNet) and DenseNet architecture. Top-1 error rates (\%) are shown. Dynamic denotes the proposed regularization method. HP: hyper-parameters. The best result under each case is bold.}
\label{table1}
\begin{tabular}{l|c|c|c}
\hline
Network Architecture                                                                                                & Params                & Regularization                                 & \begin{tabular}[c]{@{}c@{}}Top-1  \\ Error \end{tabular} \\ \hline
\multirow{4}{*}{PyramidNet-110-a48}                                                                                 & \multirow{4}{*}{1.8M} & Baseline \cite{han2017deep}                    & 23.40                                                          \\ \cline{3-4} 
                                                                                                                    &                       & ShakeDrop \cite{yamada2018shakedrop}           & 21.60                                                          \\ \cline{3-4} 
                                                                                                                    &                       & DropBlock \cite{ghiasi2018dropblock}           & 21.50                                                          \\ \cline{3-4} 
                                                                                                                    &                       & Dynamic (ours)                                 & \textbf{21.32}                                                 \\ \hline
\multirow{4}{*}{PyramidNet-26-a84}                                                                                  & \multirow{4}{*}{0.9M} & Baseline \cite{han2017deep}                    & 26.30                                                          \\ \cline{3-4} 
                                                                                                                    &                       & ShakeDrop \cite{yamada2018shakedrop}           & 31.83                                                          \\ \cline{3-4} 
                                                                                                                    &                       & DropBlock \cite{ghiasi2018dropblock}           & 23.88                                                          \\ \cline{3-4} 
                                                                                                                    &                       & Dynamic (ours)                                 & \textbf{23.83}                                                 \\ \hline
\multirow{4}{*}{PyramidNet-26-a200}                                                                                 & \multirow{4}{*}{3.8M} & Baseline \cite{han2017deep}                    & 22.53                                                          \\ \cline{3-4} 
                                                                                                                    &                       & ShakeDrop \cite{yamada2018shakedrop}           & 26.11                                                          \\ \cline{3-4} 
                                                                                                                    &                       & DropBlock \cite{ghiasi2018dropblock}           & 21.22                                                          \\ \cline{3-4} 
                                                                                                                    &                       & Dynamic (ours)                                 & \textbf{20.34}                                                 \\ \hline \hline
\multirow{4}{*}{\begin{tabular}[c]{@{}l@{}}DenseNet-BC-100-k12 \\ (default HP) \end{tabular}} & \multirow{4}{*}{0.8M} & Baseline \cite{han2017deep}                    & 22.26                                                          \\ \cline{3-4} 
                                                                                                                    &                       & ShakeDrop \cite{han2017deep} (pL=0.5)          & 25.29                                                          \\ \cline{3-4} 
                                                                                                                    &                       & DropBlock \cite{ghiasi2018dropblock} (kp=0.9)  & 23.57                                                          \\ \cline{3-4} 
                                                                                                                    &                       & Dynamic (ours)                      & \textbf{20.59}                                                 \\ \hline
\multirow{2}{*}{\begin{tabular}[c]{@{}l@{}}DenseNet-BC-100-k12 \\ (optimized HP)\end{tabular}}     & \multirow{2}{*}{0.8M} & ShakeDrop \cite{han2017deep} (pL=0.9)          & 21.41                                                          \\ \cline{3-4} 
                                                                                                                    &                       & DropBlock \cite{ghiasi2018dropblock} (kp=0.95) & 21.20 \\ \hline
\end{tabular}
\end{table}

\section{Experimental Results}\label{Experiment} 

In this section, we evaluate the proposed dynamic regularization on the classification benchmark: CIFAR100 \cite{krizhevsky2009learning} and ImageNet \cite{deng2009imagenet}, in comparison with three state-of-the-art approaches: Shake-Shake \cite{gastaldi2017shake}, ShakeDrop \cite{yamada2018shakedrop}, and DropBlock \cite{ghiasi2018dropblock}. Then we conduct ablation studies to compare the fixed or linear-increment scheme of the regularization strength, and discuss the effectiveness of the Gaussian filter and the random noise.

\subsection{Implementation Details}

\subsubsection{CIFAR100}
The following settings are used throughout the experiments. We set the training epoch to \(300\) and the batch size to \(128\). The learning rate was initialized to \(0.1\) for the 2-branch architecture \cite{yamada2018shakedrop}, and \(0.2\) for the 3-branch architecture \cite{gastaldi2017shake} and DenseNet \cite{huang2017densely}. We used the cosine learning schedule to gradually reduce the learning rate to \(0\). The weight decay and momentum was set to 0.0001 and 0.9, respectively. PyramidNet \cite{han2017deep}, ResNeXt \cite{xie2017aggregated}, and DenseNet \cite{huang2017densely} were used as baselines. We employed the standard translation, flipping \cite{krizhevsky2012imagenet} and Cutout \cite{devries2017improved} as the data augmentation scheme. Therefore, the regularizer is the only factor to affect experiments. All experimental results are presented by the average of 3 runs at the 300-th epoch.

\subsubsection{ImageNet}
ResNet-18 was trained on ImageNet-1k \cite{krizhevsky2012imagenet} with 120 epochs. The learning rate was initialized to \(0.1\). Other settings were the same as those in CIFAR100. We reported the single-crop testing results.

\subsubsection{Regularizer}
For the dynamic regularization, we set the initial dynamic factor \(s_{0}=0\) and \(A=0.5\). We used \(\Delta s=0.0003\) for the 2-branch architecture and DenseNet, and \(\Delta s=0.00025\) for the 3-branch architecture. The length of the Gaussian filter was \(501\). In ShakeDrop \cite{yamada2018shakedrop} and Shake-Shake \cite{gastaldi2017shake}, the default hyper-parameters were employed. In DropBlock \cite{ghiasi2018dropblock}, we used the default hyper-parameters in their paper, i.e., the \textit{keep\_prob} and \textit{block\_size} were set to 0.9 and 7, respectively. To prevent high regularization strength in DenseNet, we increased the value of \(p_{L}\) of ShakeDrop from 0.5 to 0.9 and increased the value of \textit{keep\_prob} of DropBlock to 0.95. To prevent underfitting of ImageNet, we applied a small \(\Delta s=5\times 10^{-7}\) in the dynamic regularization and 0.99 \textit{keep\_prob} in DropBlock.

\begin{table}[t]
\centering \caption{Comparison of regularization methods on CIFAR100 in the 3-branch architecture (i.e., ResNeXt). Top-1 error rates (\%) are shown. The best result under each case is bold.}
\label{table2}
\begin{tabular}{l|c|c|c}
\hline
Network Architecture              & Params                 & Regularization                       & Top-1 Error (\%) \\ \hline \hline
\multirow{4}{*}{ResNeXt-26-2x32d} & \multirow{4}{*}{2.9M}  & Baseline \cite{xie2017aggregated}          & 22.95            \\ \cline{3-4} 
                                  &                        & Shake-Shake \cite{gastaldi2017shake} & 21.45            \\ \cline{3-4} 
                                  &                        & DropBlock \cite{ghiasi2018dropblock} & 21.20            \\ \cline{3-4} 
                                  &                        & Dynamic (ours)                       & \textbf{20.91}   \\ \hline
\multirow{4}{*}{ResNeXt-26-2x64d} & \multirow{4}{*}{11.7M} & Baseline \cite{xie2017aggregated}          & 20.59            \\ \cline{3-4} 
                                  &                        & Shake-Shake \cite{gastaldi2017shake} & 19.19            \\ \cline{3-4} 
                                  &                        & DropBlock \cite{ghiasi2018dropblock} & 19.26            \\ \cline{3-4} 
                                  &                        & Dynamic (ours)                       & \textbf{18.76}   \\ \hline
\end{tabular}
\end{table}

\subsection{Comparison with State-of-the-Art Regularization Methods}

\subsubsection{2-branch architecture}
We start with comparing the proposed dynamic regularization with ShakeDrop \cite{yamada2018shakedrop} and DropBlock \cite{ghiasi2018dropblock} in the 2-branch architecture on CIFAR100. Following the ShakeDrop, we used PyramidNet \cite{han2017deep} as our baseline (denoted as Baseline in Table \ref{table1}) and chose different architectures including: 1) PyramidNet-110-a48 (i.e., the network has a depth of 110 layers and a widening factor of 48) which is a deep and narrow network, 2) PyramidNet-26-a84 which is a shallow network, and 3) PyramidNet-26-a200 which is a shallow and wide network.

The first three entries of Table \ref{table1} are the results of PyramidNet. From Table \ref{table1}, it can be observed that our dynamic regularization outperforms the counterparts of ShakeDrop and DropBlock in various architectures. The error rates of ShakeDrop are even worse than those of Baseline in the shallow architectures, i.e., PyramidNet-26-a84 and PyramidNet-26-a200, which means ShakeDrop with a fixed regularization strength fails in this case. This issue comes from stochastic depth \cite{huang2016deep}, where stochastic depth is designed for deep networks. With a linearly-increased dropping rate, DropBlock gains lower error rates than Baseline. However, the predefined schedule of dropping rate is inferior to our dynamic schedule. Regardless of the depth of networks, the dynamic regularization method obtains a consistent improvement.

\subsubsection{3-branch architecture}
For the 3-branch architecture, we compare the dynamic regularization with Shake-Shake \cite{gastaldi2017shake} and DropBlock \cite{ghiasi2018dropblock} in ResNeXt-26-2x32d (i.e., the network has a depth of 26 layers and 2 residual branches, and the first residual block has a width of 32 channels) and ResNeXt-26-2x64d. The results are shown in Table \ref{table2}. We can see that the error rates of dynamic regularization are lower than those of Shake-Shake and DropBlock. The results from Tables \ref{table1} and \ref{table2} show that our dynamic regularization can adapt to various network architectures. Our method can decrease the errors by more than 2\% on average in comparison with Baseline. 

\subsubsection{Densely-connected architecture}
Moreover, we evaluate the regularization methods in DenseNet-BC-100-k12 (i.e., the network uses bottleneck layers and compression with a depth of 100 layers and a growth rate of 12 \cite{huang2017densely}). The results are shown in the bottom of Table \ref{table1}. Default HP means that all regularizers employed the same hyper-parameters as the ones in PyramidNet. Optimized HP means we adjusted the hyper-parameters in terms of the regularization strength for DenseNet. With the default HP, ShakeDrop and DropBlock damaged the performance of Baseline. We found that the training errors of the two methods were much higher than the testing errors, which means the model underfitted data due to the high regularization strength. With the Optimized HP, we decreased the regularization strengths. We set a larger \textit{keep\_rate} for DropBlock and a larger \(p_{L}\) for ShakeDrop, so the performance was optimized accordingly. On the contrary, without adjusting the hyper-parameters, our dynamic regularization was stable and reduced the Top-1 error by 1.67\% (from 22.26\% to 20.59\%). 

\begin{figure} [t]
\centering 
\includegraphics[width=8.8cm]{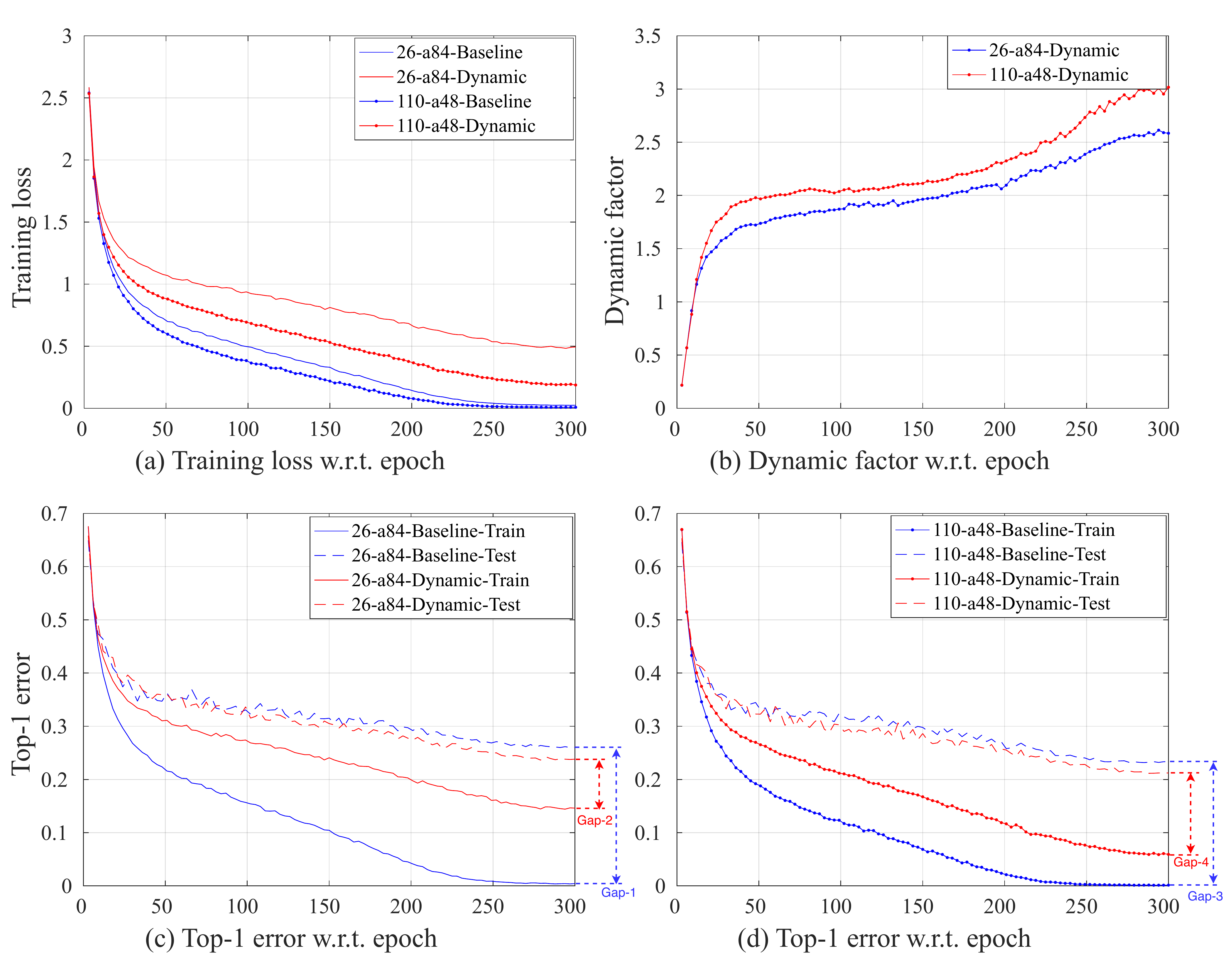}\\ 
\caption{Illustration of the training loss, dynamic factor, and Top-1 error with respect to epoch for PyramidNet. Gap stands for the difference between training and testing errors. \textit{Zoom in the figure for better viewing}.}\label{Fig5} 
\end{figure}

\begin{table}[t]
\centering \caption{Comparison of regularization methods on ImageNet, with single-crop testing. Top-1 error rates (\%) are shown. The best result is bold.}
\label{table3}
\begin{tabular}{l|c|c|c}
\hline
Network Architecture       & Params                 & Regularization                       & Top-1 Error (\%) \\ \hline \hline
\multirow{4}{*}{ResNet-18} & \multirow{4}{*}{11.7M} & Baseline \cite{he2016deep}           & 29.05            \\ \cline{3-4} 
                           &                        & ShakeDrop \cite{yamada2018shakedrop} & -                \\ \cline{3-4} 
                           &                        & DropBlock \cite{ghiasi2018dropblock} & 29.06            \\ \cline{3-4} 
                           &                        & Dynamic (ours)                       & \textbf{28.82}   \\ \hline
\end{tabular}
\end{table}

\subsubsection{Results on ImageNet}

We evaluate ResNet-18 with the dynamic regularization on ImageNet. The classification results are shown in Table \ref{table3}. Due to a large amount of data trained by a light model, ResNet-18 Baseline underfitted the training data, leading to worse performance when strong regularizers were used. ShakeDrop cannot converge on such a shallow network, so we did not report it. DropBlock also cannot work well, even though we reduced the regularization strength (i.e., the value of \textit{keep\_rate} was set to 0.99). Our dynamic regularization performed well in this situation and produced the best result (28.82\%).

\subsection{Ablation Study and Discussion}

\subsubsection{Effectiveness of dynamic regularization}
Fig. \ref{Fig5} shows the training loss, dynamic factor, and Top-1 error with respect to the epoch in the two networks, i.e., PyramidNet-26-a84 and PyramidNet-110-a48. As shown in Fig. \ref{Fig5} (a), one property of the dynamic regularization is that it prevents the training loss from a rapid descent. In other words, the networks are not easy to fit the training data by rote. Fig. \ref{Fig5} (b) illustrates that the dynamic factors of two networks gradually increase throughout training. Instead of using a predefined scheduling function in \cite{zoph2018learning}, our dynamic scheduling is self-adaptive according to the backward difference of the training loss. Another important property of the dynamic regularization is that a low regularization strength is generated for a light model (e.g., PyramidNet-26-a84), and a high strength is for a heavy model (e.g., PyramidNet-110-a48). Figs. \ref{Fig5} (c) and (d) show the networks with dynamic regularization could narrow the gap between the training and testing errors (from Gap-1 to Gap-2 for PyramidNet-26-a84 and from Gap-3 to Gap-4 for PyramidNet-110-a48, respectively) and achieve lower testing errors than the Baselines.

\subsubsection{Schedules of the regularization strength}
In \cite{zoph2018learning,ghiasi2018dropblock}, the regularization strength is adjusted by a linear-increment schedule, where ScheduledDropPath is used to linearly increase the probability of dropped path (that can also be considered as the regularization strength) in training. Besides, the fixed regularization schedule is commonly used in previous methods \cite{larsson2016fractalnet,huang2016deep,gastaldi2017shake,yamada2018shakedrop}. We used PyramidNet-26-a84 as a backbone to compare different regularization schedules.

Table \ref{table4} illustrates six different configurations of the regularization strength. `Fix-\(x\)' means the dynamic factor is fixed to \(x\) and `Linear-\(x\)' means the dynamic factor is linearly scheduled from \(0\) to \(x\) over the course of training steps. `Fix-2' and `Linear-3' achieve the best results in fixed and linear schedules, respectively. Compared with them, the dynamic setting with 23.83\% error rate achieved the best performance, which shows the effectiveness of our dynamic regularization schedule.

\begin{table}[t]
\centering \caption{Comparison of regularization schedules.}
\label{table4}
\begin{tabular}{l|c|l|c}
\hline
\multicolumn{1}{c|}{\multirow{2}{*}{PyramidNet-26-a84}} & \multirow{2}{*}{\begin{tabular}[c]{@{}c@{}}Top-1 \\ Error(\%)\end{tabular}} & \multicolumn{1}{c|}{\multirow{2}{*}{PyramidNet-26-a84}} & \multirow{2}{*}{\begin{tabular}[c]{@{}c@{}}Top-1\\ Error(\%)\end{tabular}} \\
\multicolumn{1}{c|}{} &  & \multicolumn{1}{c|}{} &  \\ \hline \hline
Fix-1 & 25.45 & Linear-1 & 25.76 \\ \hline
Fix-2 & 24.75 & Linear-2 & 25.09 \\ \hline
Fix-3 & 25.52 & Linear-3 & 24.28 \\ \hline
Fix-4 & 30.52 & Linear-4 & 25.80 \\ \hline \hline
Baseline & 26.30 & Dynamic & \textbf{23.83} \\ \hline
\end{tabular}
\end{table}

\subsubsection{Random noise}
As mentioned in Section \ref{Dynamic}, the range of the random noise involved in our dynamic regularization, i.e., $R$, is designed to grow from bottom Res-Blocks to top Res-Blocks linearly. To evaluate this setting, we performed the dynamic regularization with uniform \(R\) and linearly growing \(R\) in PyramidNet-26-a84. From the 2nd and 3rd entries of Table \ref{table5}, we can see the model with uniform \(R\) is inferior to the model with a linearly growing \(R\) (25.28\% v.s. 23.83\%).

\subsubsection{Gaussian Filtering}
In the process of updating the dynamic factor, we employed a Gaussian filter to remove the instant change of the training loss in a mini-batch mode. That is, we refer to the Eq. (\ref{eqn12}) rather than the Eq. (\ref{eqn11}) to update the dynamic factor. To study the effectiveness of the Gaussian filter, we conducted comparative experiments between the Dynamic with and without the Gaussian filter. The last two entries of Table \ref{table5} show that if we remove the Gaussian filter, the error rate increases by 1.38\%, which validates that the Gaussian filter also plays an important role in dynamic regularization.

\begin{table}[t]
\centering \caption{Effectiveness of linearly growing \(R\) and Gaussian filtering.}
\label{table5}
\begin{tabular}{l|c}
\hline
PyramidNet-26-a84          & \begin{tabular}[c]{@{}l@{}}Top-1 Error (\%)\end{tabular} \\ \hline \hline
Baseline        & 26.30                                                 \\ \hline
Dynamic-Uniform \(R\)       & 25.28                                                 \\ \hline
Dynamic-Linear growth \(R\) & \textbf{23.83}                                                 \\ \hline \hline
Dynamic-No filter       & 25.21                                                 \\ \hline
Dynamic-Gaussian filter & \textbf{23.83}                                                 \\ \hline
\end{tabular}
\end{table}

\section{Conclusion}\label{Conclusion} 

In this paper, we have presented a dynamic schedule to adjust the regularization strength to fit various network architectures and the training process. Our dynamic regularization is self-adaptive in accordance with the change of the training loss. It produces a low regularization strength for light network architectures and high regularization strength for heavy ones. Furthermore, the strength is self-paced grown to avoid overfitting. Experimental results demonstrate that the proposed dynamic regularization outperforms state-of-the-art ShakeDrop, Shake-Shake, and DropBlock regularization methods. In future, we will investigate the potential of  the dynamic regularization in data augmentation and Dropout-based methods.


%

\ifCLASSOPTIONcaptionsoff
  \newpage
\fi



%

%
\bibliographystyle{IEEEtran}
\bibliography{myref}

\begin{thebibliography}{10}
\providecommand{\url}[1]{#1}
\csname url@samestyle\endcsname
\providecommand{\newblock}{\relax}
\providecommand{\bibinfo}[2]{#2}
\providecommand{\BIBentrySTDinterwordspacing}{\spaceskip=0pt\relax}
\providecommand{\BIBentryALTinterwordstretchfactor}{4}
\providecommand{\BIBentryALTinterwordspacing}{\spaceskip=\fontdimen2\font plus
\BIBentryALTinterwordstretchfactor\fontdimen3\font minus
  \fontdimen4\font\relax}
\providecommand{\BIBforeignlanguage}[2]{{%
\expandafter\ifx\csname l@#1\endcsname\relax
\typeout{** WARNING: IEEEtran.bst: No hyphenation pattern has been}%
\typeout{** loaded for the language `#1'. Using the pattern for}%
\typeout{** the default language instead.}%
\else
\language=\csname l@#1\endcsname
\fi
#2}}
\providecommand{\BIBdecl}{\relax}
\BIBdecl

\bibitem{krizhevsky2012imagenet}
A.~Krizhevsky, I.~Sutskever, and G.~E. Hinton, ``Imagenet classification with
  deep convolutional neural networks,'' in \emph{Advances in Neural Information
  Processing Systems}, 2012, pp. 1097--1105.

\bibitem{he2016deep}
K.~He, X.~Zhang, S.~Ren, and J.~Sun, ``Deep residual learning for image
  recognition,'' in \emph{Proceedings of the IEEE Conference on Computer Vision
  and Pattern Recognition}, 2016, pp. 770--778.

\bibitem{zhao2019object}
Z.-Q. Zhao, P.~Zheng, S.-t. Xu, and X.~Wu, ``Object detection with deep
  learning: a review,'' \emph{IEEE Transactions on Neural Networks and Learning
  Systems}, 2019.

\bibitem{huang2017densely}
G.~Huang, Z.~Liu, L.~Van Der~Maaten, and K.~Q. Weinberger, ``Densely connected
  convolutional networks,'' in \emph{Proceedings of the IEEE Conference on
  Computer Vision and Pattern Recognition}, 2017, pp. 4700--4708.

\bibitem{xie2017aggregated}
S.~Xie, R.~Girshick, P.~Doll{\'a}r, Z.~Tu, and K.~He, ``Aggregated residual
  transformations for deep neural networks,'' in \emph{Proceedings of the IEEE
  Conference on Computer Vision and Pattern Recognition}, 2017, pp. 1492--1500.

\bibitem{han2017deep}
D.~Han, J.~Kim, and J.~Kim, ``Deep pyramidal residual networks,'' in
  \emph{Proceedings of the IEEE Conference on Computer Vision and Pattern
  Recognition}, 2017, pp. 5927--5935.

\bibitem{ioffe2015batch}
S.~Ioffe and C.~Szegedy, ``Batch normalization: Accelerating deep network
  training by reducing internal covariate shift,'' in \emph{International
  Conference on Machine Learning}, 2015, pp. 448--456.

\bibitem{santurkar2018does}
S.~Santurkar, D.~Tsipras, A.~Ilyas, and A.~Madry, ``How does batch
  normalization help optimization?'' in \emph{Advances in Neural Information
  Processing Systems}, 2018, pp. 2483--2493.

\bibitem{srivastava2014dropout}
N.~Srivastava, G.~Hinton, A.~Krizhevsky, I.~Sutskever, and R.~Salakhutdinov,
  ``Dropout: a simple way to prevent neural networks from overfitting,''
  \emph{The Journal of Machine Learning Research}, vol.~15, no.~1, pp.
  1929--1958, 2014.

\bibitem{ghiasi2018dropblock}
G.~Ghiasi, T.-Y. Lin, and Q.~V. Le, ``Dropblock: A regularization method for
  convolutional networks,'' in \emph{Advances in Neural Information Processing
  Systems}, 2018, pp. 10\,727--10\,737.

\bibitem{gastaldi2017shake}
X.~Gastaldi, ``Shake-shake regularization,'' \emph{CoRR}, vol. abs/1705.07485,
  2017.

\bibitem{yamada2018shakedrop}
Y.~Yamada, M.~Iwamura, T.~Akiba, and K.~Kise, ``Shakedrop regularization for
  deep residual learning,'' \emph{IEEE Access}, vol.~7, pp. 186\,126--186\,136,
  2019.

\bibitem{huang2016deep}
G.~Huang, Y.~Sun, Z.~Liu, D.~Sedra, and K.~Q. Weinberger, ``Deep networks with
  stochastic depth,'' in \emph{European Conference on Computer Vision}.\hskip
  1em plus 0.5em minus 0.4em\relax Springer, 2016, pp. 646--661.

\bibitem{bengio2009curriculum}
Y.~Bengio, J.~Louradour, R.~Collobert, and J.~Weston, ``Curriculum learning,''
  in \emph{Proceedings of the Annual International Conference on Machine
  Learning}.\hskip 1em plus 0.5em minus 0.4em\relax ACM, 2009, pp. 41--48.

\bibitem{simonyan2014very}
K.~Simonyan and A.~Zisserman, ``Very deep convolutional networks for
  large-scale image recognition,'' in \emph{International Conference on
  Learning Representations}, 2014.

\bibitem{szegedy2015going}
C.~Szegedy, W.~Liu, Y.~Jia, P.~Sermanet, S.~Reed, D.~Anguelov, D.~Erhan,
  V.~Vanhoucke, and A.~Rabinovich, ``Going deeper with convolutions,'' in
  \emph{Proceedings of the IEEE Conference on Computer Vision and Pattern
  Recognition}, 2015, pp. 1--9.

\bibitem{zagoruyko2016wide}
S.~Zagoruyko and N.~Komodakis, ``Wide residual networks,'' in \emph{British
  Machine Vision Conference}, 2016.

\bibitem{szegedy2017inception}
C.~Szegedy, S.~Ioffe, V.~Vanhoucke, and A.~A. Alemi, ``Inception-v4,
  inception-resnet and the impact of residual connections on learning,'' in
  \emph{AAAI Conference on Artificial Intelligence}, 2017.

\bibitem{devries2017improved}
T.~DeVries and G.~W. Taylor, ``Improved regularization of convolutional neural
  networks with cutout,'' \emph{CoRR}, vol. abs/1708.04552, 2017.

\bibitem{larsson2016fractalnet}
G.~Larsson, M.~Maire, and G.~Shakhnarovich, ``Fractalnet: Ultra-deep neural
  networks without residuals,'' in \emph{International Conference on Learning
  Representations}, 2017.

\bibitem{morerio2017curriculum}
P.~Morerio, J.~Cavazza, R.~Volpi, R.~Vidal, and V.~Murino, ``Curriculum
  dropout,'' in \emph{Proceedings of the IEEE International Conference on
  Computer Vision}, 2017, pp. 3544--3552.

\bibitem{goodfellow2013maxout}
I.~J. Goodfellow, D.~Warde-Farley, M.~Mirza, A.~Courville, and Y.~Bengio,
  ``Maxout networks,'' in \emph{International Conference on Machine Learning},
  2013.

\bibitem{shen2017continuous}
X.~Shen, X.~Tian, T.~Liu, F.~Xu, and D.~Tao, ``Continuous dropout,'' \emph{IEEE
  Transactions on Neural Networks and Learning Systems}, vol.~29, no.~9, pp.
  3926--3937, 2017.

\bibitem{de2003regularization}
G.~De~Nicolao and G.~Ferrari-Trecate, ``Regularization networks for inverse
  problems: A state-space approach,'' \emph{Automatica}, vol.~39, no.~4, pp.
  669--676, 2003.

\bibitem{xie2016disturblabel}
L.~Xie, J.~Wang, Z.~Wei, M.~Wang, and Q.~Tian, ``Disturblabel: Regularizing cnn
  on the loss layer,'' in \emph{Proceedings of the IEEE Conference on Computer
  Vision and Pattern Recognition}, 2016, pp. 4753--4762.

\bibitem{kumar2010self}
M.~P. Kumar, B.~Packer, and D.~Koller, ``Self-paced learning for latent
  variable models,'' in \emph{Advances in Neural Information Processing
  Systems}, 2010, pp. 1189--1197.

\bibitem{zoph2018learning}
B.~Zoph, V.~Vasudevan, J.~Shlens, and Q.~V. Le, ``Learning transferable
  architectures for scalable image recognition,'' in \emph{Proceedings of the
  IEEE Conference on Computer Vision and Pattern Recognition}, 2018, pp.
  8697--8710.

\bibitem{krizhevsky2009learning}
A.~Krizhevsky, G.~Hinton \emph{et~al.}, ``Learning multiple layers of features
  from tiny images,'' Citeseer, Tech. Rep., 2009.

\bibitem{deng2009imagenet}
J.~Deng, W.~Dong, R.~Socher, L.-J. Li, K.~Li, and L.~Fei-Fei, ``Imagenet: A
  large-scale hierarchical image database,'' in \emph{Proceedings of the IEEE
  Conference on Computer Vision and Pattern Recognition}, 2009, pp. 248--255.

\end{thebibliography}




\end{document}